\documentclass[11pt]{article}
\usepackage{amscd,amssymb,stmaryrd}

\usepackage{graphicx}
\usepackage{subcaption}
\usepackage[utf8]{inputenc}
\usepackage[export]{adjustbox}
\usepackage{wrapfig}

\usepackage{amsthm}
\usepackage{bbm}

\usepackage{color}
\usepackage{mathtools}
\usepackage{diagbox} 
\usepackage{hyperref}
\usepackage[english]{babel}
\usepackage{booktabs}

\usepackage{float}
\usepackage[linesnumbered,ruled,vlined, ruled,vlined]{algorithm2e}

\theoremstyle{definition}

\usepackage{tikz}
\usepackage{pifont}% 

\usepackage{diagbox}
\usetikzlibrary{decorations.pathreplacing,calligraphy}
\usepackage{pgfplots}
\pgfplotsset{compat=1.11}
\usepgfplotslibrary{groupplots}

\usetikzlibrary{shapes.misc}

\tikzset{cross/.style={cross out, draw=black, minimum size=2*(#1-\pgflinewidth), inner sep=0pt, outer sep=0pt},
cross/.default={1pt}}

\usepackage[linesnumbered,ruled,vlined, ruled,vlined]{algorithm2e}
\usepackage{amsmath}
\usepackage{lineno}
\usepackage{subcaption}
\usepackage{tikz}
\usepackage{xcolor}
\usepgfplotslibrary{fillbetween}
\usepackage{float}
\usepackage{pgfplots}
\pgfplotsset{compat=newest} % This command is used to ensure that you are using the latest features of the pgfplots package.

\title{DeepONet as a Multi-Operator Extrapolation Model: \\Distributed  Pretraining with Physics-Informed Fine-Tuning}
\author{Zecheng Zhang\footnote{Department of Mathematics, Florida State University, Tallahassee, FL 32304, USA. (Email: zecheng.zhang.math@gmail.com)}, 
Christian Moya\footnote{Department of Mathematics, Purdue University, West Lafayette, IN 47907, USA. (Email: cmoyacal@purdue.edu)},
Lu Lu\footnote{Department of Statistics and Data Science, Yale University, New Haven, CT 06511, USA. (Email: lu.lu@yale.edu)}, Guang Lin\footnote{Department of Mathematics and Mechanical Engineering, Purdue University, West Lafayette, IN 47907, USA. (Email: guanglin@purdue.edu)},
Hayden Schaeffer \footnote{Department of Mathematics, UCLA, Los Angeles, CA 90095, USA. (Email: hayden@math.ucla.edu)}
}
\date{}

\begin{document}
\maketitle
 
% ====================================
% abstract
% ====================================
\begin{abstract}
We propose a novel fine-tuning method to achieve multi-operator learning through training a distributed neural operator with diverse function data and then zero-shot fine-tuning the neural network using physics-informed losses for downstream tasks. 
Operator learning effectively approximates solution operators for PDEs and various PDE-related problems, yet it often struggles to generalize to new tasks. 
To address this, we investigate fine-tuning a pretrained model, while carefully selecting an initialization that enables rapid adaptation to new tasks with minimal data. 
Our approach combines distributed learning to integrate data from various operators in pre-training, while physics-informed methods enable zero-shot fine-tuning, minimizing the reliance on downstream data.
We investigate standard fine-tuning and Low-Rank Adaptation fine-tuning, applying both to train complex nonlinear target operators that are difficult to learn only using random initialization. Through comprehensive numerical examples, we demonstrate the advantages of our approach, showcasing significant improvements in accuracy. Our findings provide a robust framework for advancing multi-operator learning and highlight the potential of transfer learning techniques in this domain.
\end{abstract}

% =====================
% section: introduction
% =====================
\section{Introduction} \label{sec:intro}
In recent years, neural operators~\cite{chen1995universal, li2020fourier,zhang2022belnet,lu2022comprehensive,jiao2021one} and deep operator learning~\cite{chen1995universal,lu2021learning,jin2022mionet} have emerged as powerful tools for approximating mappings between function spaces, particularly in the context of solving partial differential equations (PDEs)~\cite{lin2023b,moya2023bayesian,zhang2024bayesian,jiao2024solving} and ordinary differential equations (ODEs)~\cite{moya2024conformalized, lin2023learning}. These techniques hold great promise for various applications, including physics simulations and engineering problems~\cite{moya2023deeponet, moya2023approximating, mao2023ppdonet, zhu2023fourier,sahin2024deep,mao2024disk2planet,lee2024efficient,yin2024dimon,jiang2024fourier}. However, traditional approaches often face significant challenges, primarily due to the requirement for extensive datasets~\cite{lu2022multifidelity,di2023neural} and their inability to effectively handle heterogeneous input spaces~\cite{lin2021operator}. As operators become increasingly complex, the need for more efficient learning methodologies grows~\cite{cai2021deepm,mao2021deepm}.

To address the data limitations inherent in operator learning, physics-informed DeepONets~\cite{wang2021learning,goswami2022physics,goswami2023physics, leung2022nh} have been developed. These models are designed to incorporate physical laws that govern the behavior of PDEs and ODEs, thereby reducing the amount of training data needed while enhancing model interpretability. Despite their advantages, training physics-informed DeepONets presents its own difficulties, particularly in identifying appropriate initial training points. The complexity of the underlying physical systems often complicates the selection of these points, which can hinder the overall training process.

Apart from data limitations, another key challenge in neural operator learning lies in handling extrapolation. This involves leveraging a pretrained model to predict test samples that exhibit properties distinct from the training examples. For instance, in \cite{zhu2023reliable}, the authors address the challenge of predicting downstream tasks with variations in input functions or function domains, and they propose methods to quantify differences between input function distributions.
To tackle generalization and extrapolation for new operators—such as solution operators associated with unseen PDEs — MOL approaches \cite{sun2024towards, yang2023prompting, zhang2024modno, liu2023prose, liu2024prose, jollie2024time, sun2024lemon, subramanian2024towards, ye2024pdeformer, mccabe2023multiple} have been developed. MOL aims to use a unified framework to learn multiple operators simultaneously. By incorporating structures that encode operator-specific information, such as symbolic encodings \cite{liu2023prose, sun2024towards, sun2024lemon, jollie2024time,liu2024prosefd}, MOL may have the ability to extrapolate to different input distributions and unseen operators.

However, most MOL frameworks are computationally intensive, requiring extensive datasets, and still face challenges in generalizing to PDEs with entirely new physical properties. To address downstream tasks of arbitrary forms, fine-tuning with streaming data \cite{sun2024lemon} can be an effective strategy. The success of this technique relies on a well-initialized model and the quality of the downstream task data. In scientific applications related to solving PDE; however, selecting a suitable initialization or identifying a model that closely resembles the new tasks is challenging \cite{schaeffer2017learning}. Additionally, sufficient downstream data may not be available.
In this work, we employ distributed learning algorithms to achieve a robust initialization adaptable to downstream tasks. Furthermore, we use physics-informed training techniques to enable zero-shot tuning, allowing for effective adjustment without the need for downstream data.

Federated learning \cite{moya2022fed, zhang2024federated}, Deep Distributed Neural Operators (D2NO) \cite{zhang2024d2no}, and Multi-Operator Learning with Distributed Neural Operator (MODNO) \cite{zhang2024modno} were introduced to address the challenges posed by heterogeneous multiscale input spaces and multi-operator learning (MOL). In particular, D2NO and MODNO facilitate distributed learning by enabling the construction of distinct operators tailored to specific input spaces. This modular framework not only enhances flexibility but also leverages the collective contributions of all operators within a D2NO, effectively capturing a wide range of responses. These empirical properties make it an ideal foundation for training operators in complex scenarios. Some recent advancements in scientific machine learning indicate that training with a diverse dataset could produce a pretrained model that adapts quickly to downstream tasks \cite{sun2024lemon, bodnar2024aurora}. For example, in \cite{sun2024lemon}, the authors demonstrate that including data from diverse PDE solution operators improves the model’s capacity for extrapolation and fine-tuning. In this work, we utilize the capabilities of D2NO/MODNO to combine data from different operators, generating a robust pretrained model that is adaptable for new tasks.

In this paper, we propose the use of fine-tuning as a form of transfer learning to improve operator training. While fine-tuning has been extensively utilized in natural language processing (NLP)~\cite{dodge2020fine}, its application in the domain of operator learning remains very limited~\cite{zhu2023reliable}. We classify fine-tuning into two categories: full fine-tuning, where all parameters of the pretrained model are updated during the training process, and parameter-efficient fine-tuning~\cite{hu2021lora,hu2023llm,ding2023parameter}, which updates only a small subset of parameters while keeping the majority of the model fixed. Our objective is to leverage these fine-tuning techniques to enhance the training of complex nonlinear target operators that are often challenging to learn from random initialization, whether due to their complexity or insufficient data. 
By leveraging pretrained physics-informed DeepONets and D2NOs as starting points, we aim to establish a zero-shot fine-tuning approach and a robust pretraining model tailored for the downstream extrapolation tuning process.

We summarize the contributions of our paper below.
\begin{enumerate}
    \item \textit{Enhanced Physics-Informed Operator Learning:} We present a fine-tuning method that extends pretrained DeepONets to achieve more accurate zero-shot operator fine-tuning extrapolation. This approach leverages the physical principles embedded in both the pretrained and target operators, enabling models that not only outperform those trained from random initialization but also produce predictions consistent with fundamental physical laws. Remarkably, this is achieved without requiring additional supervised fine-tuning data, facilitating \textit{physics-informed zero-shot extrapolation.}
    \item \textit{Fine-Tuning Deep Distributed Neural Operators:} We introduce a fine-tuning strategy tailored for D2NOs, showcasing how these operators can use prior knowledge from diverse multiple pretrained models. This approach enhances flexibility and adaptability, enabling the model to accurately approximate complex target operators across diverse scenarios through physics-informed zero-shot extrapolation.
    \item \textit{Low Rank Adaptation for Efficient Fine-Tuning:} We implement LoRA techniques for both physics-informed DeepONets and D2NOs, significantly reducing the number of parameters that need to be trained during fine-tuning. This improves computational efficiency while preserving the core strengths of the original pretrained models, making them more viable for application in resource-constrained environments. The potential trade-off comes from a reduction in the model's expressiveness in certain operator contexts.
\end{enumerate}
We organize the rest of the paper as follows. Section~\ref{sec:review} reviews DeepONets, physics-informed DeepONets, and D2NOs. Section~\ref{sec:method} describes the proposed fine-tuning methods, including LoRA, for both DeepONet and D2NO. In Section~\ref{sec:numerical-examples}, we provide three examples that demonstrate the advantages of the proposed method compared to training new operators from random initialization. Finally, Section~\ref{sec:conclusion} concludes the paper and discusses our future work.  
% ===============
% section: review
% ===============
\section{Background} \label{sec:review}
This section reviews the related operator networks: DeepONets, Physics-Informed (PI) DeepONets, and D2NO/MODNO.
% ====================
% subsection: DeepONet
% ====================
\subsection{Deep Operator Networks}  \label{subsec:deeponet}
 DeepONets~\cite{lu2021learning, chen1995universal}, are a novel class of neural network architectures designed to learn operators that map functions to functions. They are particularly advantageous in applications involving PDEs, where the operator is often complex and difficult to describe analytically. Formally, let \( \mathcal{U} \) be a space of functions and let \( \mathcal{V} \) be the corresponding output space. A typical operator \( G : \mathcal{U} \to \mathcal{V} \) maps a function \( u \in \mathcal{U} \) to another function \( v \in \mathcal{V} \).

The architecture of a DeepONet consists of two main networks: the \textit{branch network} and the \textit{trunk network}. The branch networks take as input a discretization $\hat{u}$ of the input function \( u \), discretized using \(m\) sensors (sampled values). 
In contrast, the trunk network captures the spatial or temporal domain of the output function, receiving a set of input coordinates \( x \in \mathbb{R}^d \) and learning to map these coordinates to the corresponding output values.
The outputs of the branch and trunk networks are combined (via a dot product) to yield the output of the DeepONet with trainable parameters denoted as $\theta$ as follows:
\begin{align}
    G[u](x) \approx G_{\theta}[\hat{u}](x) = \sum_{k=1}^{K} p_k(\hat{u}) b_k(x), u\in \mathcal{U},
    \label{eqn_approx}
\end{align}
where \( G[u](x) \) is the predicted output function evaluated at the coordinates \( x \), and $p_k$ denotes the branch nets while $b_k(x)$ is the trunk basis net. The training procedure for DeepONets involves using pairs of input-output function data to learn the underlying operator. Given a dataset of triplets \( (u_i, x_i, v_i) \) where \( u_i \) is a function from \( \mathcal{U} \) and \( v_i = G[u_i](x_i) \) is the corresponding output evaluated as \(x_i\), the network parameters \( \theta \in \mathbb{R}^p \) are optimized by minimizing a loss function defined as:
\[
\mathcal{L}(\theta) = \frac{1}{N} \sum_{i=1}^{N} \left\| v_i - G_{\theta}[u_i](x_i) \right\|^2,
\]
where \( G_{\theta}[u_i](y_i) \) is the output of the DeepONet for the input function \( u_i \) at the coordinates \( x_i \), and \( N \) is the total number of training samples.
Once trained, DeepONets can efficiently evaluate the learned operator on new input functions, making them a powerful tool for approximating complex function mappings in various applications~\cite{deng2022approximation}.
% ===========================
% subsec: Deep Distributed NO
% ===========================
\subsection{Deep Distributed Neural Operators } \label{subsec:D2NO}
D2NOs ~\cite{zhang2024d2no, zhang2024modno} are an advanced framework developed to efficiently learn mappings between function spaces in distributed computing environments. D2NO was initially proposed to address input function spaces with heterogeneous properties by processing functions with similar characteristics locally. 
This framework was later extended to MODNO (Multi-Operator learning with Distributed Neural Operators) \cite{zhang2024modno} to tackle the challenge of multi-operator learning by managing distinct output functions associated with different operators in a localized manner. 

Consider we have \( C \) sub-datasets, \( D_1, D_2, \ldots, D_C \), where each sub-dataset corresponds to a unique set of data. In D2NO setting, one sub-dataset contains the input functions presenting similar properties; while in MODNO settings, it is the data associated with one operator. Each local client \( c \) manages its own local dataset $D_c$, and in the D2NO setting, the local loss function for client \( c \) can be defined as:
\[
L_c(\alpha_c; \beta) = \sum_{i=1}^{N_c} \| G[u_{c,i}](\cdot) - G_{\alpha_c, \beta}[\hat{u}_{c,i}](\cdot) \|^2, \ \ u_{c,i}\in D_c
\]
where \( N_c \) is the number of input functions in dataset \( D_c \) and \( \alpha_c \) and \( \beta \) are the local and global parameters for the branch networks to learn dedicated function encoding and trunk networks to learn the shared output function basis, respectively. 
Notably, with a slight abuse of notation, we will always use $\alpha_c$ to denote the local parameters and $\beta$ to denote the globally shared parameters.
For MODNO multi-operator learning setting, the $c$-th local loss for operator $G_c$ can be defined as,
% \begin{align*}
%     L_c(\alpha_c; \beta) = \sum_{i = 1}^{N_u} \|G_c(\hat{u}_{c, i})(x_c) - \sum_{k = 1}^K p_k(\hat{u}_{c, i}; \beta_k) b_{k,c}(x_c; \alpha_{k, c}) \|^2,
% \end{align*}
\begin{align*}
    L_c(\alpha_c; \beta) = \sum_{i = 1}^{N_u} \| G_c[u_{c,i}](\cdot) - G_{\alpha_c, \beta}[\hat{u}_{c,i}](\cdot) \|^2,
\end{align*}
where $u_{c, i}$ is the input function of the operator $G_c$, \( \alpha_c \) represents the local parameters for the trunk networks, which learn the basis for the output functions of different operators, while \( \beta \) denotes the shared parameters used to encode the input functions.
The global loss function  which is used to update the shared parameters $\beta$ across all clients using all data is then defined as:
\[
L(\beta; \alpha) = \sum_{c=1}^{C} L_c(\alpha_c; \beta).
\]

The training procedure for D2NO/MODNO involves the following steps:

\begin{enumerate}
    \item \textit{Initialization:} Initialize the weights and trainable parameters for each branch network and the trunk network.
    
    \item \textit{Local Updates:} For each client \( c \):
    \begin{itemize}
        \item Compute the local loss \( L_c(\alpha_c; \beta) \) using its dedicated dataset \( D_c \).
        \item Update the local parameters $\alpha_c$:
        \[
        \alpha_c \leftarrow \alpha_c - \eta_c \nabla_{\alpha_c} L_c(\alpha_c; \beta),
        \]
        where \( \eta_c \) is the learning rate for client \( c \).
    \end{itemize}

    \item \textit{Global Synchronization:} After local updates, the centralized server aggregates the local parameters and updates the shared network:
    \[
    \beta \leftarrow \beta - \eta \nabla_{\beta} L(\beta; \alpha),
    \]
    where \( \eta \) is the learning rate for the shared networks.

    \item \textit{Iteration:} Repeat the local updates and global synchronization for a predetermined number of iterations or until convergence.
\end{enumerate}

This distributed approach allows D2NO/MODNO to efficiently learn operators while accommodating heterogeneous input functions and MOL ultimately achieving improved predictive accuracy and computational efficiency in various numerical experiments.

Finally, to facilitate the use of D2NO/MODNO for fine-tuning, we average the weights of the local networks. This process effectively conserves information from a diverse set of responses of an operator across multiple input spaces. The pretrained weights are computed as follows:
\[
\alpha = \frac{1}{C} \sum_{c=1}^C \alpha_c.
\]
This merging process not only captures the variability of responses but also provides a beneficial warm start for fine-tuning, yielding a robust initialization that enhances the model’s adaptability to target operators. 

\subsection{Physics-Informed Learning}
Physics-Informed Neural Networks (PINN) is a data-free neural network approach for solving PDEs~\cite{lu2021deepxde,karniadakis2021physics}. This technique can be applied to DeepONet, enabling it to approximate operators. For instance, consider a one-dimensional PDE \( u_t - u_{xx} = f \) defined on \([0, 1]\) with periodic boundary conditions. 
Here, we aim to learn the operator that maps the initial condition (IC) to the solution at a later time. 
Let \( u_i \) denote the $i$-th discretized input functions in the training dataset,
we then denote the $j$-th point in its output function's domain as $(x_{ij},t_{ij})$, i.e. the quantity $G[u_i](x_{ij},t_{ij})$, which is used for training the neural operator.
We then solve the following minimization problem: 
\begin{align*}
   \min_{\theta} & \quad \omega_1 \sum_{i = 1}^{N_u}\sum_{j = 1}^{N_x} |\mathcal{AD}_{t}\{G_{\theta}[u_i]({x}_{ij}, t_{ij}) \}-\mathcal{AD}_{xx}\{G_{\theta}[u_i]({x}_{ij}, t_{ij}) \} - f({x}_{ij}, t_{ij})|^2 \\
    &\quad + \omega_2 \sum_{i = 1}^{N_u}\sum_{j = 1}^{N_{ic}}|G_{\theta}[u_i](x_{ij}, 0) - u_i(x_{ij})|^2\\
    &\quad+ \omega_3  \sum_{i = 1}^{N_u}\sum_{j = 1}^{N_{bc}} |G_{\theta}[u_i](0, t_{ij}) - G_{\theta}[u_i](1, t_{ij}) |^2.
\end{align*}
Here, \( \mathcal{AD}_{xx} \) denotes the second-order derivative with respect to the network's spatial input \( x \) (the trunk input), while \( \mathcal{AD}_t \) represents the first-order derivative with respect to the network's temporal input \( t \) (also a trunk input); both derivatives are implemented using automatic differentiation. The variable $\omega_i$ represents the weights. The last term enforces periodic boundary conditions.

Since optimizing DeepONet with PI training only needs the IC and boundary conditions (BC) and does not require data, PI offers a supervised approach to approximate the operator. However, directly applying PI to a randomly initialized neural operator can lead to performance challenges for many problems, such as slow convergence.
In this work, we demonstrate that with a MODNO/D2NO-initialized neural operator, PI can optimize the network in a zero-shot setting (without data) and significantly enhance prediction accuracy.

% ================
% section: methods
% ================
\section{Methodology} \label{sec:method}
In this section, we introduce our proposed methods, including adopting a MODNO/D2NO pretrained and averaged neural operator as well as PI fine-tuning.

% =============================
% sub-section: Full Fine-tuning
% =============================
\subsection{Full Fine-Tuning of DeepONets}  \label{subsec:full_finetuning}
Full fine-tuning is a conventional approach utilized to adapt pretrained neural networks for specific tasks by updating \textit{all model parameters} during the training process. This method allows the model to leverage the knowledge acquired during pretraining while also incorporating task-specific information, resulting in improved performance on target datasets. The main idea behind full fine-tuning is to refine the entire set of model parameters, enabling the model to capture complex features relevant to the new task.

In the context of a neural network layer with weight matrix \( W \in \mathbb{R}^{m \times n} \), full fine-tuning involves directly updating the weights during the training process. The updated weight matrix can be represented as:
\[
\tilde{W} = W + \Delta W,
\]
where \( \Delta W \) represents the changes applied to the weight matrix during fine-tuning. This approach allows the model to effectively adjust to the nuances of the new data, improving its ability to approximate complex operators.

To implement full fine-tuning in the linear layers of the branch and trunk networks in a DeepONet or merged D2NO, we first identify the linear trainable layers within these networks. Let \( \mathcal{B} \) and \( \mathcal{T} \) denote the branch and trunk networks, respectively, containing linear layers with weights \( W_{\mathcal{B}} \) and \( W_{\mathcal{T}} \). During the fine-tuning process, both \( W_{\mathcal{B}} \) and \( W_{\mathcal{T}} \) are updated as follows:
\[
\tilde{W}_{\mathcal{B}} = W_{\mathcal{B}} + \Delta W_{\mathcal{B}},
\]
\[
\tilde{W}_{\mathcal{T}} = W_{\mathcal{T}} + \Delta W_{\mathcal{T}}.
\]
In this context, \( \Delta W_{\mathcal{B}} \) and \( \Delta W_{\mathcal{T}} \) are computed based on the gradients obtained from the physics-informed loss function with respect to the target operators. This process allows the model to effectively adapt its weights to the specific characteristics of the target data, thereby enhancing its performance. During training, we optimize all parameters of \( W_{\mathcal{B}} \) and \( W_{\mathcal{T}} \) simultaneously, enabling the fine-tuned DeepONet to learn the required adjustments for the target functions. This approach leads to improved convergence and performance on complex operator approximations compared to training from random initialization. Once the full fine-tuning is completed, we evaluate the model on new input functions and target operators, capitalizing on the pretrained knowledge encapsulated in the updated parameters. This comprehensive adaptation ensures that the model can generalize effectively while accurately approximating the desired target operators.
% ===============================
% subsection: Low Rank Adaptation
% ===============================
\subsection{Low Rank Adaptation Finetuning of DeepONets}  \label{subsec:lora} 
LoRA~\cite{hu2021lora} is a technique designed to efficiently fine-tune pretrained neural networks by adding low-rank parameter matrices to the existing model parameters. This approach enables the model to adapt to new tasks with significantly fewer trainable parameters, reducing computational costs while maintaining performance. The main idea behind LoRA is to decompose the weight updates of the network into low-rank matrices, allowing for the capture of task-specific features without the need for full fine-tuning of all model parameters. In the context of a neural network layer with weight matrix \( W \in \mathbb{R}^{m \times n} \), LoRA introduces two low-rank matrices, \( A \in \mathbb{R}^{m \times r} \) and \( B \in \mathbb{R}^{r \times n} \), where \( r \) is much smaller than \( m \) and \( n \). During fine-tuning, instead of updating \( W \) directly, the model learns the updates as follows:
\[
\tilde{W} = W + A B,
\]
where \( \tilde{W} \) is the adapted weight matrix used during inference. This structure allows the model to leverage the pretrained knowledge while efficiently learning task-specific adjustments, leading to enhanced adaptability with reduced computational overhead.

To apply LoRA fine-tuning to the linear layers of the branch and trunk networks in a DeepONet, we first identify the linear transformations in these networks. Let \( \mathcal{B} \) and \( \mathcal{T} \) represent the branch and trunk networks, respectively, which contain linear layers with weights \( W_{\mathcal{B}} \) and \( W_{\mathcal{T}} \). Instead of fine-tuning \( W_{\mathcal{B}} \) and \( W_{\mathcal{T}} \) directly, we introduce low-rank matrices \( A_{\mathcal{B}}, B_{\mathcal{B}} \) for the branch network and \( A_{\mathcal{T}}, B_{\mathcal{T}} \) for the trunk network. The updated weights for these layers during fine-tuning are then expressed as:
\[
\tilde{W}_{\mathcal{B}} = W_{\mathcal{B}} + A_{\mathcal{B}} B_{\mathcal{B}},
\]
\[
\tilde{W}_{\mathcal{T}} = W_{\mathcal{T}} + A_{\mathcal{T}} B_{\mathcal{T}}.
\]

During training, we only optimize the parameters of the low-rank matrices \( A_{\mathcal{B}}, B_{\mathcal{B}}, A_{\mathcal{T}}, B_{\mathcal{T}} \), while keeping \( W_{\mathcal{B}} \) and \( W_{\mathcal{T}} \) frozen. This approach enables the DeepONet to adaptively learn additional operators for the target functions with a fraction of the parameters typically required, allowing for faster convergence and improved efficiency in training. Once the LoRA parameters are trained, we can evaluate the finetuned on new input functions and target operators while still maintaining the benefits of the pretrained knowledge encapsulated in the frozen layers.

\subsection{MODNO/D2NO Pretraining and PI Fine Tuning}
A strong initialization is crucial for downstream tasks, where a ``good'' initialization means the model can quickly adapt to the new task with minimal data. In operator learning, a well-suited initialization would be a neural operator trained on data from an operator with behavior similar to the downstream task. However, there is limited research comparing different operators for this purpose. Even when the pretraining and downstream operators exhibit similar behavior, differences in the input function distributions can hinder performance if the pretrained operator is directly used for initialization.

\begin{figure}
    \centering
    \includegraphics[scale = 0.12]{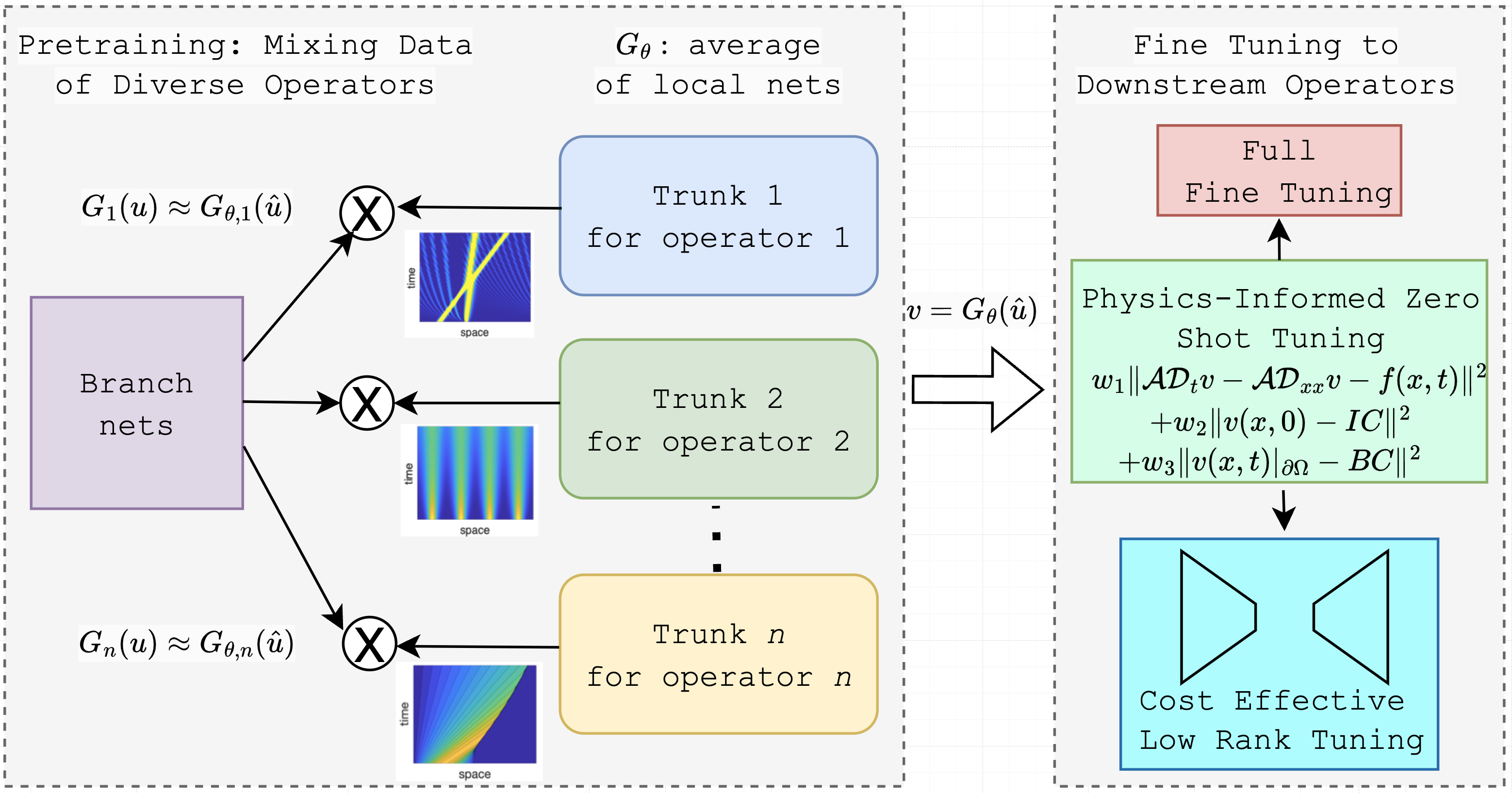}
    \caption{Methodology demonstration for downstream PDE $u_t - u_{xx} = f$ with initial condition (IC) and Dirchlet boundary conditions (BC). $\mathcal{AD}$ denotes the auto-differentiaiton of the modern machine learning software, $\bigotimes$ denotes the inner product.}
    \label{fig_demon}
\end{figure}

Recent work in multi-operator learning foundation models offers valuable insights for achieving robust initialization by training on data from multiple operators. The key idea is to train a model on a diverse set of operators, enabling it to adapt quickly to new tasks. However, since DeepONet is designed for single-operator learning, it approximates only one operator at a time, limiting its use with multi-operator data. To address this limitation, we adopt the MODNO/D2NO algorithm. Specifically, since training requires mixing data from different operators to achieve MOL, we will use the MODNO approach. In MODNO pretraining, we train a shared data encoder (branch network) for the input function and use separate trunk networks to represent output functions for different operators. After pretraining, we average the models to create an effective initialization.
As D2NO was introduced before MODNO, we will use D2NO in our naming conventions in the numerical example sections.
To adapt to downstream tasks, we will use PI tuning, enabling zero-shot adaptation for extrapolation tasks involving new operators. The numerical results in the next section validate the effectiveness of the proposed method. We present a workflow of our methods in Figure \ref{fig_demon}.

% ===========================
% section: numerical-examples
% ===========================
\section{Numerical Examples} \label{sec:numerical-examples}
This section presents four numerical examples that illustrate the advantages of our proposed fine-tuning methods over fine-tuning/retraining a neural network trained with one operator data and training new neural networks from random initialization to approximate operators. In our notation, PI-LoRA refers to physics-informed LoRA fine-tuning in a zero-shot setting, while PI-Full denotes the standard zero-shot fine-tuning approach.

\subsection{Example~1: Extrapolation of Burgers'-Type Equations}
In the first example, we focus on the extrapolation of Burgers'-type equations. Specifically, we investigate the mapping from the initial condition to the solution at the terminal simulation time for the following PDE:
\begin{align}
    u_t + \nu \left(\frac{u^2}{2}\right )_x = \mu u_{xx},  \ \ t\in[0, T],\ \  x\in[0, 2\pi],
    \label{eqn_vburgers}
\end{align}
with periodic boundary conditions. The initial conditions are generated using the Gaussian mixture model described in Section~\ref{sec_example1_setting}. To obtain models that will serve as the initialization for fine-tuning to the target operator, we consider operators from three different equations:~(1) Equation \ref{eqn_vburgers} with parameters $\mu = 0.02, \nu = 0.5, T = 1$; (2) Equation \ref{eqn_vburgers} with $\mu = 0.03, \nu = 0.5, T = 1$; and (3) equation \ref{eqn_vburgers} with $\mu = 0.05, \nu = 0.5, T = 0.5$. The target operator corresponds to the mapping from the initial condition to the solution of equation \ref{eqn_vburgers} with $\mu = 0.01, \nu = 0.5, T = 1$.

\begin{figure}[H]
    \centering
    \includegraphics[scale = 0.45]{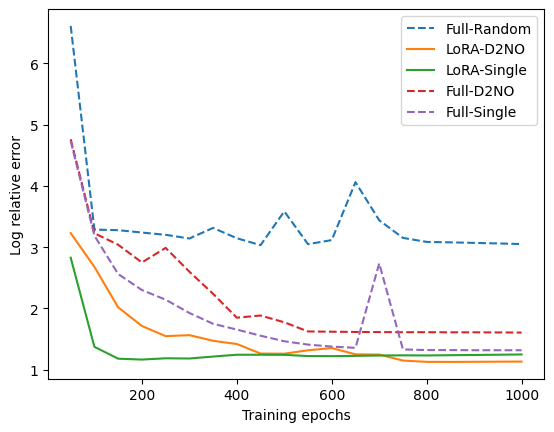}
    \includegraphics[scale = 0.45]{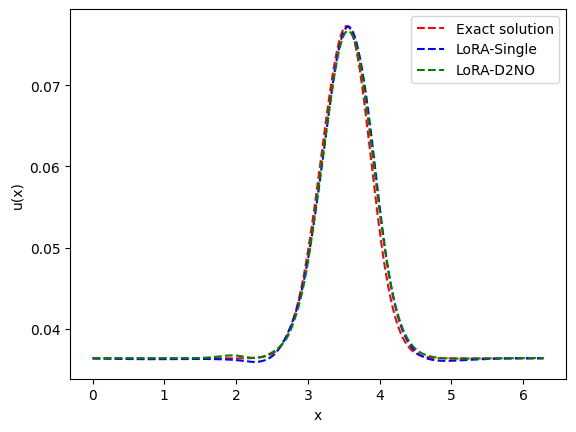}
    \caption{Left: Relative error (in log scale) decay with respect to training epochs. Three full tuning curves are dashed lines, and two LoRA training curves are solid lines. The final relative errors are presented in Table \ref{table_vburgers_results}. Right: A demonstration of the predictions.}
    \label{fig_eg1_err}
\end{figure}

We compared three pretraining models: (1) pretraining using the D2NO/MODNO algorithm, (2) pretraining with data from a single operator with $\mu = 0.05, \nu = 0.5$, and (3) random initialization without pretraining. For each setting, we conducted experiments independently with different random seeds, running each experiment 10 times, and presenting the average results in Table \ref{table_vburgers_results} and training error decay trajectory in Figure \ref{fig_eg1_err}.

\begin{table}[h!]
    \centering
    \begin{tabular}{|c|c|c|c|}
        \hline
        \diagbox{Fine-tune}{Pretrain} & D2NO & $\mu = 0.05, \nu = 0.5$ & Random  \\
        \hline
        PI-LoRA & $3.11\%$ & $3.49\%$  & - \\
        \hline
        PI-Full & $4.99\%$ & $3.74\%$ & $21.14\%$\\
        \hline
    \end{tabular}
    \caption{Average relative errors across all testing samples. The errors represent the mean of 8 independent runs with different random seeds. The standard deviations for various experimental settings are as follows: 0.15 for LoRA D2NO pretraining, 0.31 for LoRA single-operator pretraining, and 4.02 for random initialization.  All numerical experiments follow the same experimental settings, including training epochs, loss weights, learning rate scheduler, and other parameters. The relative error decay with respective to the training epochs is presented in Figure \ref{fig_eg1_err}. }
    \label{table_vburgers_results}
\end{table}

\subsubsection{Analysis of Results}
Firstly, We can observe from Figure \ref{fig_eg1_err} that all model trainings have been stabilized.
As seen in the last column of Table \ref{table_vburgers_results}, a randomly initialized model fails to provide accurate predictions in a zero-shot setting, underscoring the importance of fine-tuning. Second, when fine-tuning a model trained with data from an operator exhibiting similar properties, the performance shows a significant improvement over random initialization or training from scratch. However, a search algorithm for an operator with properties similar to the target operator remains underexplored in the research. Therefore, training an initialization using data from multiple operators may provide a better starting point for downstream tasks. The table shows that the proposed D2NO/MODNO, which averages data from multiple operators, provides a robust initialization, enabling the fine-tuned model to achieve optimal performance. This highlights the effectiveness of using federated algorithms for model pretraining, as demonstrated in~\cite{mollaali2024conformalized} for an engineering problem. Additionally, we observe an improvement in accuracy when using PI-LoRA for this experiment. As outlined in Section \ref{sec_example1_setting}, PI-LoRA tunes far fewer parameters than PI-Full, with a parameter count of 32.9K compared to 65.2K. This indicates a substantial reduction in computational device memory usage.

\subsubsection{Pretraining and Fine-Tuning Details}
\label{sec_example1_setting}
To generate the pretraining data, we create the initial conditions using normal distributions with random means and variances. The means and standard deviations are uniformly sampled from the intervals \([2\pi, 4\pi]\) and \([0.3, 1]\), respectively. For the testing data, or the initial conditions for the testing operator, we follow the same process but shift the distribution by \(0.01\) to increase the difficulty of the extrapolation. In the PI fine-tuning process, we use 17 temporal points and 40 spatial points, uniformly sampled, to construct the equation loss. For sampling the initial condition, we use 26 points along with 200 input-output function pairs. No additional data is used during the fine-tuning process.

The network consists of 10 basis functions. Each branch network has a fully connected structure with dimensions \(N_i \times 100 \rightarrow 100 \times 1\), where \(N_i = 13\) represents the input dimension or discretization mesh size. For the LoRA implementation in the branch networks, we set the rank to 10. The trunk network follows a fully connected structure with layers sized \(N_i \times 100 \rightarrow 100 \times 100 \rightarrow 100 \times 100 \rightarrow 100 \times 100 \rightarrow 100 \times 100 \rightarrow 100 \times 10\). For the LoRA implementation in the trunk network, we set the rank to 10 for all layers except the last, where the rank is set to 4. This network structure is used in all subsequent experiments.

\subsection{Example 2: Extrapolation of the Porous Media Equations}
\label{sec_pm_example}
In this example, we consider the porous media equations of different orders, specifically:
\begin{align}
    u_t -(u_{xx})^m = f, \quad x\in[0, 2],~t\in[0, 0.01].
    \label{eqn_nl_porous_media}
\end{align}
We investigate the D2NO pretrained model, which is trained using data from three operators that map the initial condition to the solution at later times: equation (\ref{eqn_nl_porous_media}) with (1) \(m = 1\) and \(f(x) = \frac{1}{5}\sin(2\pi x)\), (2) \(m = 3\) and \(f(x) = \frac{1}{5}\cos(2\pi x)\), and (3) \(m = 4\) and \(f(x) = \frac{1}{10}\sin(2\pi x)\). We test the target operator with \(m = 2\) and \(f(x) = \frac{1}{5}\sin(2\pi x)\).

\begin{table}[h!]
    \centering
    \begin{tabular}{|c|c|c|c|c|}
        \hline
        \diagbox{Fine-tune}{Pretrain} & D2NO & $m = 1$ porous media & Random  \\
        \hline
        PI-LoRA & $5.49\%$ & $6.14\%$  & -  \\
        \hline
        PI-Full & $6.57\%$ & $8.51\%$ & $12.57\%$ \\
        \hline
    \end{tabular}
    \caption{Average relative errors for all testing samples in the porous media equation example. The errors represent the average of 10 independent runs with different random seeds. The standard deviations for various experimental settings are as follows: 0.26 for PI-LoRA D2NO pretraining, 0.25 for PI-LoRA single-operator pretraining, and 2.50 for random initialization. All numerical experiments use the same settings (training epochs, loss weights, etc.). Notably, the training and testing operators do not share the same input function distributions, as described in Section \ref{sec_porous_training_setting}. }
    \label{table_pm_results}
\end{table}

% do not delete the comments below
% single pretrain: 6.6,6.2,5.95,5.95,6.19,6.81,5.75,6.36,6.13,6.8, std 0.25
% random: 7.95, 12.45, 14.31, 14.28, 14.3, 14.22, 14.29, 8.82, 14.28, 14.28, std 2.50
% d2no: 5.38 5.24 5.38 5.56 5.41 6.07 6.41 6.66 5.23 5.65 std 0.26.

\subsubsection{Analysis of Results}
From Table \ref{table_pm_results}, we observe that using PINN-DeepONet to directly solve the target operators results in a relatively high error of $12.57\%$. However, when we use a pretrained model that closely approximates the target, as proposed, the error significantly decreases to $6.14\%$, as shown in the third column. Selecting an operator with similar properties can be challenging, though. To address this, we applied the MODNO/D2NO-based method, which pretrains an averaged model. This approach further reduces the error to $5.49\%$, demonstrating that the proposed methods provide a robust initialization for downstream tasks.

\subsubsection{Pretraining and Fine-Tuning Details}
\label{sec_porous_training_setting}
In this example, the training and testing input functions (initial conditions of the PDEs) are generated by different functions. Specifically, the input functions for the training and testing operators are defined as:
$u(x; \mathbf{w}) = w_1\sin(\pi x) + w_2\sin(2\pi x) + w_3\sin(4\pi x) + w_4\sin(6\pi x) + w_5\cos(\pi x) + w_6\cos(2\pi x) + w_7\cos(4\pi x) + w_8\cos(6\pi x) + w_9,$ 
where \(w_i \sim \mathcal{U}(-2, 2)\) for \(i = 1, \dots, 8\) and \(w_9 \sim \mathcal{U}(0.1, 2)\). In the PI fine-tuning process, we use a \(17 \times 40\) temporal-spatial discretization and study 250 input-output function pairs.

\subsection{Example 3:~Extrapolation of the Diffusion-Reaction Equations}
\label{sec_example_dr}
In this example, we consider the diffusion-reaction equations, given by:
\begin{align} \label{eqn_dr}
u_t - u_{xx} + g(u) = f(x), \quad x \in [0, 2], \, t \in [0, 0.01],
\end{align} 
with Dirichlet boundary conditions \(u(0) = u(2) = 1\). The D2NO pretrained model is based on three operators that map the initial condition to the solutions at later times: (1) \(g(u) = -u\) and \(f(x) = \exp{x}\), (2) \(g(u) = -5u^2\) and \(f(x) = x^2\), and (3) \(g(u) = -u(1-u)\) and \(f(x) = \sin(x)\). The target operator corresponds to the solution operator of the equation \(g(u) = -(u-0.5)(u-1)\) and \(f(x) = \cos(\pi x)\). The results are presented in Table \ref{table_rd_results}.

\begin{table}[H]
    \centering
    \begin{tabular}{|c|c|c|c|c|}
        \hline
        \diagbox{Fine-tune}{Pretrain} & D2NO & $g(u) = -5u^2$, $f(x) = x^2$ & Random  \\
        \hline
        PI-LoRA & $3.24\%$ & $3.41\%$  & -  \\
        \hline
        PI-Full & $3.43\%$ & $5.14\%$ & $4.66\%$ \\
        \hline
    \end{tabular}
    \caption{Average relative errors for all testing samples in the diffusion-reaction equation example. The errors represent the average of 10 independent runs with different random seeds.
    The standard deviations for various experimental settings are as follows: 0.04 for LoRA D2NO pretraining, 0.24 for LoRA single-operator pretraining, and 0.46 for random initialization.
    All numerical experiments use the same settings (training epochs, loss weights, etc.). Notably, the training and testing operators do not share the same input function distributions, as detailed in Section \ref{sec_rd_training_setting}.}
    \label{table_rd_results}
\end{table}

% d2no: 0.04
%  single 0.24
% rdm: 0.46

\subsubsection{Analysis of Results}
Similar to the first two examples, the model pretrained using the proposed MODNO/D2NO approach outperforms both single-operator data pretraining and random initialization. Furthermore, PI-LoRA fine-tuning outperforms PI-Full in both time/memory efficiency and prediction accuracy. Specifically, in terms of computational time, PI-LoRA slightly outperforms PI-Full, taking 187 seconds compared to 191 seconds. However, the difference in computational memory is more significant, with PI-LoRA requiring only 32,860 parameters, while PI-Full requires 65,200 parameters.

\subsubsection{Pretraining and Fine-Tuning Details}
\label{sec_rd_training_setting}
We first present the input functions used for training and testing. Specifically, for all operators involved in pretraining and testing, the input functions are generated by the following series:
$u(x; \mathbf{w}) = w_1\sin(\pi x) + w_2\sin(2\pi x) + w_3\sin(4\pi x) + w_4\sin(6\pi x) + w_5\cos(\pi x) + w_6\cos(2\pi x) + w_7\cos(4\pi x) + w_8\cos(6\pi x) + w_9,$
where \(w_i \sim \mathcal{U}(-2, 2)\) for \(i = 1, \dots, 8\) and \(w_9 \sim \mathcal{U}(0.1, 2)\). In the physics-informed fine-tuning process, we use a \(17 \times 40\) temporal-spatial discretization and study 250 input-output function pairs. The network structures and LoRA training details remain the same as in the first experiment.

\subsection{Example 4: Extrapolation of Two Different PDE Families}
\label{sec_pm_to_rd}
\begin{figure}[H]
    \centering
    \includegraphics[scale = 0.5]{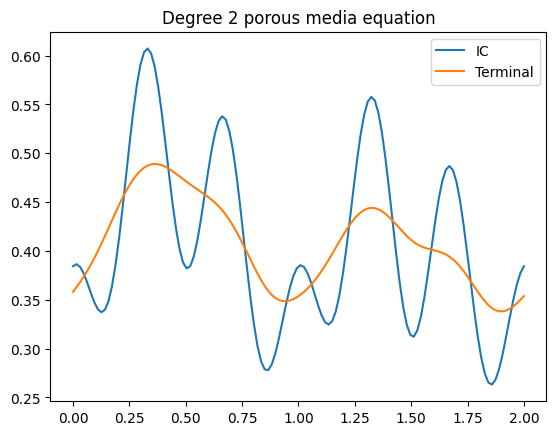}
    \includegraphics[scale = 0.5]{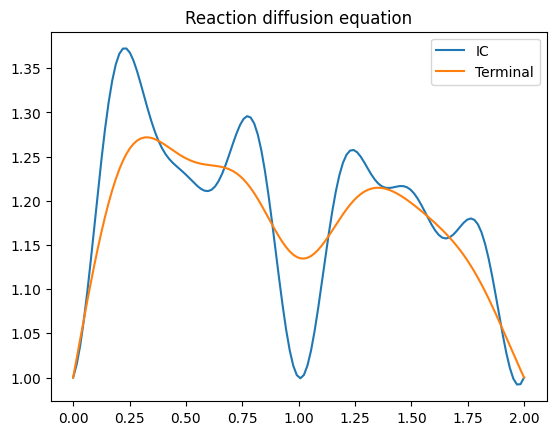}
    \caption{Demonstration for two solution operators: (1) diffusion-reaction system (\ref{eqn_dr}) with  $g(u) = -(u-0.5)(u-1)$ and $f(x) = \cos(\pi x)$, (2) Porous media system (\ref{eqn_nl_porous_media}) with degree 2 and $f(x) = \frac{1}{5} \sin(2\pi x)$. Porous media solutions are utilized in constructing the MODNO/D2NO pretraining model, while the downstream task is the reaction-diffusion system.
    We ran the numerical experiments for 8 times with different random seeds, and the standard deviations for various experimental settings are as follows: 0.04 for RD pretrained model, 0.05 for PM pretrained model, and 0.46 for random initialization.
    }
    \label{fig_comp_pm_rd}
\end{figure}

In this example, we consider the case where the pretraining operators and downstream operators belong to two different PDE families. Specifically, we pretrain using operators from three porous media equations, as outlined in Section \ref{sec_pm_example}, and then apply a diffusion-reaction type downstream operator, as discussed in Section \ref{sec_example_dr}. We present the solutions of these two operators in Figure \ref{fig_comp_pm_rd} and summarize the results in Table~\ref{table_cross_results}.
\begin{table}[h!]
    \centering
    \begin{tabular}{|c|c|c|c|c|}
        \hline
        \diagbox{Fine-tune}{Pretrain} & RD & PM & Random  \\
        \hline
        PI-LoRA & $3.24\%$ & $3.33\%$  & $4.66\%$  \\
        \hline
    \end{tabular}
    \caption{Average relative errors for all testing samples in the reaction-diffusion (RD) example from Section \ref{sec_pm_to_rd}. Notably, we study the pretraining model derived from porous media (PM) equations, which differ from the downstream target tasks (RD). The errors represent the average of 10 independent runs with different random seeds. All numerical experiments use the same settings (training epochs, loss weights, etc.)  
    }
    \label{table_cross_results}
\end{table}

% pm pretraining: 0.05

%\subsubsection{Results Analysis}

\section{Conclusion} \label{sec:conclusion}
In this work, we have presented an innovative framework for fine-tuning operator learning methodologies by integrating physics-informed DeepONets and Deep Distributed Neural Operators. By addressing the limitations of traditional operator learning approaches, we have shown that pretrained models can be effectively used to reduce data requirements and enhance the accuracy of operator approximations. Our use of fine-tuning techniques offers flexibility in training complex nonlinear target operators, enabling efficient adaptations to various scenarios. In our future work, we plan to extend this framework to scientific foundation models by combining fine-tuning with uncertainty quantification, ensuring proper adaptability in diverse contexts. Our findings support a shift towards leveraging pretrained models in multi-operator learning, promoting more efficient and interpretable machine learning solutions in complex scientific and engineering applications.

\section*{Acknowledgement}
ZZ would like to thank the U.S. Department of Energy (DOE) Office of Science Advanced Scientific Computing Research program DE-SC0025440.
GL and CM would like to thank the support of the National Science Foundation (DMS-2053746, DMS-2134209, ECCS-2328241, CBET-2347401 and OAC-2311848), and U.S.~Department of Energy (DOE) Office of Science Advanced Scientific Computing Research program DE-SC0023161, the Uncertainty Quantification for Multifidelity Operator Learning (MOLUcQ) project (Project No. 81739), and DOE–Fusion Energy Science, under grant number: DE-SC0024583. LL was supported by
the U.S. DOE Office of Advanced Scientific Computing Research under Grants No.~DE-SC0025593 and No.~DE-SC0025592, and the U.S. National Science Foundation under Grant No.~DMS-2347833. HS was supported in part by NSF DMS 2427558 and NSF DMS 2331033.

\bibliographystyle{unsrt}
\bibliography{references}

\begin{thebibliography}{10}

\bibitem{chen1995universal}
Tianping Chen and Hong Chen.
\newblock Universal approximation to nonlinear operators by neural networks with arbitrary activation functions and its application to dynamical systems.
\newblock {\em IEEE Transactions on Neural Networks}, 6(4):911--917, 1995.

\bibitem{li2020fourier}
Zongyi Li, Nikola Kovachki, Kamyar Azizzadenesheli, Burigede Liu, Kaushik Bhattacharya, Andrew Stuart, and Anima Anandkumar.
\newblock Fourier neural operator for parametric partial differential equations.
\newblock {\em arXiv preprint arXiv:2010.08895}, 2020.

\bibitem{zhang2022belnet}
Zecheng Zhang, Wing~Tat Leung, and Hayden Schaeffer.
\newblock Belnet: Basis enhanced learning, a mesh-free neural operator.
\newblock {\em arXiv preprint arXiv:2212.07336}, 2022.

\bibitem{lu2022comprehensive}
Lu~Lu, Xuhui Meng, Shengze Cai, Zhiping Mao, Somdatta Goswami, Zhongqiang Zhang, and George~Em Karniadakis.
\newblock A comprehensive and fair comparison of two neural operators (with practical extensions) based on fair data.
\newblock {\em Computer Methods in Applied Mechanics and Engineering}, 393:114778, 2022.

\bibitem{jiao2021one}
Anran Jiao, Haiyang He, Rishikesh Ranade, Jay Pathak, and Lu~Lu.
\newblock One-shot learning for solution operators of partial differential equations.
\newblock {\em arXiv preprint arXiv:2104.05512}, 2021.

\bibitem{lu2021learning}
Lu~Lu, Pengzhan Jin, Guofei Pang, Zhongqiang Zhang, and George~Em Karniadakis.
\newblock Learning nonlinear operators via deeponet based on the universal approximation theorem of operators.
\newblock {\em Nature Machine Intelligence}, 3(3):218--229, 2021.

\bibitem{jin2022mionet}
Pengzhan Jin, Shuai Meng, and Lu~Lu.
\newblock Mionet: Learning multiple-input operators via tensor product.
\newblock {\em SIAM Journal on Scientific Computing}, 44(6):A3490--A3514, 2022.

\bibitem{lin2023b}
Guang Lin, Christian Moya, and Zecheng Zhang.
\newblock B-deeponet: An enhanced bayesian deeponet for solving noisy parametric pdes using accelerated replica exchange sgld.
\newblock {\em Journal of Computational Physics}, 473:111713, 2023.

\bibitem{moya2023bayesian}
Christian Moya~Calderon and Guang Lin.
\newblock Bayesian, multifidelity operator learning for complex engineering systems-a position paper.
\newblock {\em Journal of Computing and Information Science in Engineering}, pages 1--9, 2023.

\bibitem{zhang2024bayesian}
Zecheng Zhang, Christian Moya, Wing~Tat Leung, Guang Lin, and Hayden Schaeffer.
\newblock Bayesian deep operator learning for homogenized to fine-scale maps for multiscale pde.
\newblock {\em Multiscale Modeling \& Simulation}, 22(3):956--972, 2024.

\bibitem{jiao2024solving}
Anran Jiao, Qile Yan, Jhn Harlim, and Lu~Lu.
\newblock Solving forward and inverse pde problems on unknown manifolds via physics-informed neural operators.
\newblock {\em arXiv preprint arXiv:2407.05477}, 2024.

\bibitem{moya2024conformalized}
Christian Moya, Amirhossein Mollaali, Zecheng Zhang, Lu~Lu, and Guang Lin.
\newblock Conformalized-deeponet: A distribution-free framework for uncertainty quantification in deep operator networks.
\newblock {\em arXiv preprint arXiv:2402.15406}, 2024.

\bibitem{lin2023learning}
Guang Lin, Christian Moya, and Zecheng Zhang.
\newblock Learning the dynamical response of nonlinear non-autonomous dynamical systems with deep operator neural networks.
\newblock {\em Engineering Applications of Artificial Intelligence}, 125:106689, 2023.

\bibitem{moya2023deeponet}
Christian Moya, Shiqi Zhang, Guang Lin, and Meng Yue.
\newblock Deeponet-grid-uq: A trustworthy deep operator framework for predicting the power grid’s post-fault trajectories.
\newblock {\em Neurocomputing}, 535:166--182, 2023.

\bibitem{moya2023approximating}
Christian Moya, Guang Lin, Tianqiao Zhao, and Meng Yue.
\newblock On approximating the dynamic response of synchronous generators via operator learning: A step towards building deep operator-based power grid simulators.
\newblock {\em arXiv preprint arXiv:2301.12538}, 2023.

\bibitem{mao2023ppdonet}
Shunyuan Mao, Ruobing Dong, Lu~Lu, Kwang~Moo Yi, Sifan Wang, and Paris Perdikaris.
\newblock Ppdonet: Deep operator networks for fast prediction of steady-state solutions in disk--planet systems.
\newblock {\em The Astrophysical Journal Letters}, 950(2):L12, 2023.

\bibitem{zhu2023fourier}
Min Zhu, Shihang Feng, Youzuo Lin, and Lu~Lu.
\newblock Fourier-deeponet: Fourier-enhanced deep operator networks for full waveform inversion with improved accuracy, generalizability, and robustness.
\newblock {\em Computer Methods in Applied Mechanics and Engineering}, 416:116300, 2023.

\bibitem{sahin2024deep}
Izzet Sahin, Christian Moya, Amirhossein Mollaali, Guang Lin, and Guillermo Paniagua.
\newblock Deep operator learning-based surrogate models with uncertainty quantification for optimizing internal cooling channel rib profiles.
\newblock {\em International Journal of Heat and Mass Transfer}, 219:124813, 2024.

\bibitem{mao2024disk2planet}
Shunyuan Mao, Ruobing Dong, Kwang~Moo Yi, Lu~Lu, Sifan Wang, and Paris Perdikaris.
\newblock Disk2planet: A robust and automated machine learning tool for parameter inference in disk-planet systems.
\newblock {\em arXiv preprint arXiv:2409.17228}, 2024.

\bibitem{lee2024efficient}
Jonathan~E Lee, Min Zhu, Ziqiao Xi, Kun Wang, Yanhua~O Yuan, and Lu~Lu.
\newblock Efficient and generalizable nested fourier-deeponet for three-dimensional geological carbon sequestration.
\newblock {\em arXiv preprint arXiv:2409.16572}, 2024.

\bibitem{yin2024dimon}
Minglang Yin, Nicolas Charon, Ryan Brody, Lu~Lu, Natalia Trayanova, and Mauro Maggioni.
\newblock Dimon: Learning solution operators of partial differential equations on a diffeomorphic family of domains.
\newblock {\em arXiv preprint arXiv:2402.07250}, 2024.

\bibitem{jiang2024fourier}
Zhongyi Jiang, Min Zhu, and Lu~Lu.
\newblock Fourier-mionet: Fourier-enhanced multiple-input neural operators for multiphase modeling of geological carbon sequestration.
\newblock {\em Reliability Engineering \& System Safety}, 251:110392, 2024.

\bibitem{lu2022multifidelity}
Lu~Lu, Rapha{\"e}l Pestourie, Steven~G Johnson, and Giuseppe Romano.
\newblock Multifidelity deep neural operators for efficient learning of partial differential equations with application to fast inverse design of nanoscale heat transport.
\newblock {\em Physical Review Research}, 4(2):023210, 2022.

\bibitem{di2023neural}
Patricio~Clark Di~Leoni, Lu~Lu, Charles Meneveau, George~Em Karniadakis, and Tamer~A Zaki.
\newblock Neural operator prediction of linear instability waves in high-speed boundary layers.
\newblock {\em Journal of Computational Physics}, 474:111793, 2023.

\bibitem{lin2021operator}
Chensen Lin, Zhen Li, Lu~Lu, Shengze Cai, Martin Maxey, and George~Em Karniadakis.
\newblock Operator learning for predicting multiscale bubble growth dynamics.
\newblock {\em The Journal of Chemical Physics}, 154(10), 2021.

\bibitem{cai2021deepm}
Shengze Cai, Zhicheng Wang, Lu~Lu, Tamer~A Zaki, and George~Em Karniadakis.
\newblock Deepm\&mnet: Inferring the electroconvection multiphysics fields based on operator approximation by neural networks.
\newblock {\em Journal of Computational Physics}, 436:110296, 2021.

\bibitem{mao2021deepm}
Zhiping Mao, Lu~Lu, Olaf Marxen, Tamer~A Zaki, and George~Em Karniadakis.
\newblock Deepm\&mnet for hypersonics: Predicting the coupled flow and finite-rate chemistry behind a normal shock using neural-network approximation of operators.
\newblock {\em Journal of computational physics}, 447:110698, 2021.

\bibitem{wang2021learning}
Sifan Wang, Hanwen Wang, and Paris Perdikaris.
\newblock Learning the solution operator of parametric partial differential equations with physics-informed deeponets.
\newblock {\em Science advances}, 7(40):eabi8605, 2021.

\bibitem{goswami2022physics}
Somdatta Goswami, Minglang Yin, Yue Yu, and George~Em Karniadakis.
\newblock A physics-informed variational deeponet for predicting crack path in quasi-brittle materials.
\newblock {\em Computer Methods in Applied Mechanics and Engineering}, 391:114587, 2022.

\bibitem{goswami2023physics}
Somdatta Goswami, Aniruddha Bora, Yue Yu, and George~Em Karniadakis.
\newblock Physics-informed deep neural operator networks.
\newblock In {\em Machine Learning in Modeling and Simulation: Methods and Applications}, pages 219--254. Springer, 2023.

\bibitem{leung2022nh}
Wing~Tat Leung, Guang Lin, and Zecheng Zhang.
\newblock Nh-pinn: Neural homogenization-based physics-informed neural network for multiscale problems.
\newblock {\em Journal of Computational Physics}, page 111539, 2022.

\bibitem{zhu2023reliable}
Min Zhu, Handi Zhang, Anran Jiao, George~Em Karniadakis, and Lu~Lu.
\newblock Reliable extrapolation of deep neural operators informed by physics or sparse observations.
\newblock {\em Computer Methods in Applied Mechanics and Engineering}, 412:116064, 2023.

\bibitem{sun2024towards}
Jingmin Sun, Yuxuan Liu, Zecheng Zhang, and Hayden Schaeffer.
\newblock Towards a foundation model for partial differential equations: Multi-operator learning and extrapolation.
\newblock {\em arXiv preprint arXiv:2404.12355}, 2024.

\bibitem{yang2023prompting}
Liu Yang, Tingwei Meng, Siting Liu, and Stanley~J Osher.
\newblock Prompting in-context operator learning with sensor data, equations, and natural language.
\newblock {\em arXiv preprint arXiv:2308.05061}, 2023.

\bibitem{zhang2024modno}
Zecheng Zhang.
\newblock Modno: Multi-operator learning with distributed neural operators.
\newblock {\em Computer Methods in Applied Mechanics and Engineering}, 431:117229, 2024.

\bibitem{liu2023prose}
Yuxuan Liu, Zecheng Zhang, and Hayden Schaeffer.
\newblock Prose: Predicting operators and symbolic expressions using multimodal transformers.
\newblock {\em Neural Networkks}, 180:106707, 2024.

\bibitem{liu2024prose}
Yuxuan Liu, Zecheng Zhang, and Hayden Schaeffer.
\newblock Prose: Predicting operators and symbolic expressions using multimodal transformers.
\newblock {\em arXiv preprint arXiv:2309.16816}, 2023.

\bibitem{jollie2024time}
Derek Jollie, Jingmin Sun, Zecheng Zhang, and Hayden Schaeffer.
\newblock Time-series forecasting, knowledge distillation, and refinement within a multimodal pde foundation model.
\newblock {\em arXiv preprint arXiv:2409.11609}, 2024.

\bibitem{sun2024lemon}
Jingmin Sun, Zecheng Zhang, and Hayden Schaeffer.
\newblock Lemon: Learning to learn multi-operator networks.
\newblock {\em arXiv preprint arXiv:2408.16168}, 2024.

\bibitem{subramanian2024towards}
Shashank Subramanian, Peter Harrington, Kurt Keutzer, Wahid Bhimji, Dmitriy Morozov, Michael~W Mahoney, and Amir Gholami.
\newblock Towards foundation models for scientific machine learning: Characterizing scaling and transfer behavior.
\newblock {\em Advances in Neural Information Processing Systems}, 36, 2024.

\bibitem{ye2024pdeformer}
Zhanhong Ye, Xiang Huang, Leheng Chen, Zining Liu, Bingyang Wu, Hongsheng Liu, Zidong Wang, and Bin Dong.
\newblock Pdeformer-1: A foundation model for one-dimensional partial differential equations.
\newblock {\em arXiv preprint arXiv:2407.06664}, 2024.

\bibitem{mccabe2023multiple}
Michael McCabe, Bruno R{\'e}galdo-Saint Blancard, Liam~Holden Parker, Ruben Ohana, Miles Cranmer, Alberto Bietti, Michael Eickenberg, Siavash Golkar, Geraud Krawezik, Francois Lanusse, et~al.
\newblock Multiple physics pretraining for physical surrogate models.
\newblock {\em arXiv preprint arXiv:2310.02994}, 2023.

\bibitem{liu2024prosefd}
Yuxuan Liu, Jingmin Sun, Xinjie He, Griffin Pinney, Zecheng Zhang, and Hayden Schaeffer.
\newblock Prose-fd: A multimodal pde foundation model for learning multiple operators for forecasting fluid dynamics.
\newblock {\em arXiv preprint arXiv:2409.09811}, 2024.

\bibitem{schaeffer2017learning}
Hayden Schaeffer.
\newblock Learning partial differential equations via data discovery and sparse optimization.
\newblock {\em Proceedings of the Royal Society A: Mathematical, Physical and Engineering Sciences}, 473(2197):20160446, 2017.

\bibitem{moya2022fed}
Christian Moya and Guang Lin.
\newblock Fed-deeponet: Stochastic gradient-based federated training of deep operator networks.
\newblock {\em Algorithms}, 15(9):325, 2022.

\bibitem{zhang2024federated}
Handi Zhang, Langchen Liu, and Lu~Lu.
\newblock Federated scientific machine learning for approximating functions and solving differential equations with data heterogeneity.
\newblock {\em arXiv preprint arXiv:2410.13141}, 2024.

\bibitem{zhang2024d2no}
Zecheng Zhang, Christian Moya, Lu~Lu, Guang Lin, and Hayden Schaeffer.
\newblock D2no: Efficient handling of heterogeneous input function spaces with distributed deep neural operators.
\newblock {\em Computer Methods in Applied Mechanics and Engineering}, 428:117084, 2024.

\bibitem{bodnar2024aurora}
Cristian Bodnar, Wessel~P Bruinsma, Ana Lucic, Megan Stanley, Johannes Brandstetter, Patrick Garvan, Maik Riechert, Jonathan Weyn, Haiyu Dong, Anna Vaughan, et~al.
\newblock Aurora: A foundation model of the atmosphere.
\newblock {\em arXiv preprint arXiv:2405.13063}, 2024.

\bibitem{dodge2020fine}
Jesse Dodge, Gabriel Ilharco, Roy Schwartz, Ali Farhadi, Hannaneh Hajishirzi, and Noah Smith.
\newblock Fine-tuning pretrained language models: Weight initializations, data orders, and early stopping.
\newblock {\em arXiv preprint arXiv:2002.06305}, 2020.

\bibitem{hu2021lora}
Edward~J Hu, Yelong Shen, Phillip Wallis, Zeyuan Allen-Zhu, Yuanzhi Li, Shean Wang, Lu~Wang, and Weizhu Chen.
\newblock Lora: Low-rank adaptation of large language models.
\newblock {\em arXiv preprint arXiv:2106.09685}, 2021.

\bibitem{hu2023llm}
Zhiqiang Hu, Lei Wang, Yihuai Lan, Wanyu Xu, Ee-Peng Lim, Lidong Bing, Xing Xu, Soujanya Poria, and Roy Ka-Wei Lee.
\newblock Llm-adapters: An adapter family for parameter-efficient fine-tuning of large language models.
\newblock {\em arXiv preprint arXiv:2304.01933}, 2023.

\bibitem{ding2023parameter}
Ning Ding, Yujia Qin, Guang Yang, Fuchao Wei, Zonghan Yang, Yusheng Su, Shengding Hu, Yulin Chen, Chi-Min Chan, Weize Chen, et~al.
\newblock Parameter-efficient fine-tuning of large-scale pre-trained language models.
\newblock {\em Nature Machine Intelligence}, 5(3):220--235, 2023.

\bibitem{deng2022approximation}
Beichuan Deng, Yeonjong Shin, Lu~Lu, Zhongqiang Zhang, and George~Em Karniadakis.
\newblock Approximation rates of deeponets for learning operators arising from advection--diffusion equations.
\newblock {\em Neural Networks}, 153:411--426, 2022.

\bibitem{lu2021deepxde}
Lu~Lu, Xuhui Meng, Zhiping Mao, and George~Em Karniadakis.
\newblock Deepxde: A deep learning library for solving differential equations.
\newblock {\em SIAM review}, 63(1):208--228, 2021.

\bibitem{karniadakis2021physics}
George~Em Karniadakis, Ioannis~G Kevrekidis, Lu~Lu, Paris Perdikaris, Sifan Wang, and Liu Yang.
\newblock Physics-informed machine learning.
\newblock {\em Nature Reviews Physics}, 3(6):422--440, 2021.

\bibitem{mollaali2024conformalized}
Amirhossein Mollaali, Gabriel Zufferey, Gonzalo Constante-Flores, Christian Moya, Can Li, Guang Lin, and Meng Yue.
\newblock Conformalized prediction of post-fault voltage trajectories using pre-trained and finetuned attention-driven neural operators.
\newblock {\em arXiv preprint arXiv:2410.24162}, 2024.

\end{thebibliography}
\end{document}